\documentclass[journal,comsoc]{IEEEtran}
\usepackage[lofdepth,lotdepth]{subfig}
\usepackage[T1]{fontenc}
\usepackage{tabularx,lipsum,booktabs}

\usepackage[switch]{lineno}
\usepackage{lipsum}
\usepackage{multicol}
\usepackage{blindtext,graphicx}
\usepackage{mathtools}
\usepackage{commath}
\usepackage{amsmath}
\usepackage{epsfig}
\usepackage{epstopdf}
\usepackage{comment}
\usepackage{cite}
\usepackage{listings}
\usepackage{amssymb}
\usepackage{soul}

\usepackage[breaklinks=true]{hyperref}
\usepackage{graphicx}
\usepackage[english]{babel}
\usepackage{xcolor}
\usepackage[cmintegrals]{newtxmath}
\usepackage[utf8x]{inputenc} 
\begin{document}
\title{Towards On-Device AI and Blockchain for 6G enabled Agricultural Supply-chain Management}
\author{Author 1 ......}

\author{\IEEEauthorblockN{Muhammad Zawish, Nouman Ashraf, Rafay Iqbal Ansari, ‪Steven Davy, Hassaan Khaliq Qureshi, Nauman Aslam, Syed Ali Hassan}}
\maketitle
\let\thefootnote\relax\footnotetext{M. Zawish \textbf{\textcolor{blue}{(Corresponding Author)}} and S. Davy are with Walton Institute for Information and Communication Systems Science, Waterford Institute of Technology (WIT), Waterford, Ireland\\ Emails:\{muhammad.zawish, steven.davy\}@waltoninstitute.ie\\ N. Ashraf is with Walton Institute, Ireland and Turku University of Applied Sciences, Finland  Email: \{nouman.ashraf\}@turkuamk.fi\\
R. I. Ansari and N. Aslam are with the Department of Computer and Information Sciences, Northumbria University Newcastle, UK.\\
Emails: \{rafay.ansari, nauman.aslam\}@northumbria.ac.uk\\
H. K. Qureshi and S. A. Hassan are with the Department of Electrical Engineering, SEECS, National University of Sciences and Technology, Pakistan.\\
Email: \{hassaan.khaliq, ali.hassan\}@seecs.edu.pk 
}

\maketitle
\begin{abstract}
   6G envisions artificial intelligence (AI) powered solutions for enhancing the quality-of-service (QoS) in the network and to ensure optimal utilization of resources. In this work, we propose an architecture based on the combination of unmanned aerial vehicles (UAVs), AI and blockchain for agricultural supply-chain management with the purpose of ensuring traceability, transparency, tracking inventories and contracts. We propose a solution to facilitate on-device AI by generating a roadmap of  models with various resource-accuracy trade-offs. A fully convolutional neural network (FCN) model is used for biomass estimation through images captured by the UAV. Instead of a single compressed FCN model for deployment on UAV, we motivate the idea of iterative pruning to provide multiple task-specific models with various complexities and accuracy. To alleviate the impact of flight failure in a 6G enabled dynamic UAV network, the proposed model selection strategy will assist UAVs to update the model based on the runtime resource requirements. 
\end{abstract}
\IEEEpeerreviewmaketitle

\section{Introduction}
\IEEEPARstart{T}{he} development of 6G communication has brought about several new technologies to enhance user experience by providing seamless connectivity and high quality-of-service (QoS). The technologies that have undergone development under the aegis of 5G networks include machine-to-machine (M2M) communication, internet of everything (IoE) and device-to-device (D2D) networks, to name a few \cite{sun1}. Recently, the utilization of unmanned aerial vehicles (UAVs) as access points has been explored as a viable candidate for providing on-demand services \cite{wang1} for 6G networks. The UAV base stations can provide connectivity to ground users in an ad-hoc manner, thereby supporting the conventional network infrastructure. Moreover, artificial intelligence (AI) empowered solutions have been proposed to enhance the QoS in 6G networks \cite{helin}. To this end, a significant amount of work has focused on using AI to optimise 6G enabled wireless systems, but limited attention has been paid to device level resource optimisation of AI models, specifically for use in UAVs. Thus, the AI-led 6G networks will leverage the opportunities offered by the aforementioned technologies for the development of new applications.

In this work, we propose an architecture based on UAVs empowered by AI and blockchain for agriculture supply-chain management by capturing images of the fields and estimating biomass through UAVs. We propose a system model based on a combination of blockchain and on-device AI for ensuring traceability, transparency, tracking provenance of crops by observing the state of farms, inventories and contracts in the agriculture supply-chain.  Typically, the UAV is only responsible to collect data via its sensors and transmit it to a remote cloud server for processing. Although this cloud-centric approach helps UAVs in saving energy by transmitting computation-intensive tasks, the wireless transmission of raw data introduces significant latency and security issues. However, shifting AI from cloud to the device restricts unnecessary data transfer and ensures network security as a key performance indicator (KPI) of 6G enabled wireless systems \cite{wang1,helin,tariq}. The proposed architecture is sensitive to both latency and privacy which can not be tolerated in a cloud-centric approach, thus on-device computation is the preferred approach. However, UAVs usually possess limited computational and low-power capabilities which hamper the task of on-device processing. Thereby, the computational complexity of the underlying AI algorithm plays a key role in realizing on-device processing \cite{canziani}.

Deep learning as an extension of AI has resulted in numerous prominent algorithms such as convolutional neural networks (CNNs) for image classification, segmentation etc. Recently, CNNs have caught the spotlight in a wide range of mobile vision applications such as self-driving cars, cyber-physical systems, autonomous systems and many more \cite{canziani, long}. In this work, we propose the use of a fully convolutional neural network (FCN) a type of CNN \cite{long} for semantic segmentation of biomass pixels from the image captured by the UAV. Based on the above discussion, embedding a FCN directly into a UAV is not a feasible solution due to its limited resources. To enable the on-device processing using CNNs, recent works have focused on providing a fixed compressed model using techniques such as network pruning, weight quantization, and knowledge distillation \cite{han01}. Instead of providing a single compressed model using the above techniques for on-device execution, we leverage the idea of iterative pruning to provide multiple task-specific models with various resource-accuracy trade-offs. This approach is more significant when performing on-device processing because different UAVs possess different battery timing, processing and storage capacities \cite{wang1}, thus, it is not feasible to design a model with rigid characteristics. The aim of providing models with various complexities and accuracy trade-offs is to allow UAVs to fetch the required model based on the dynamic resources during the flight.

\begin{figure*}[h]
\includegraphics[width=1\textwidth]{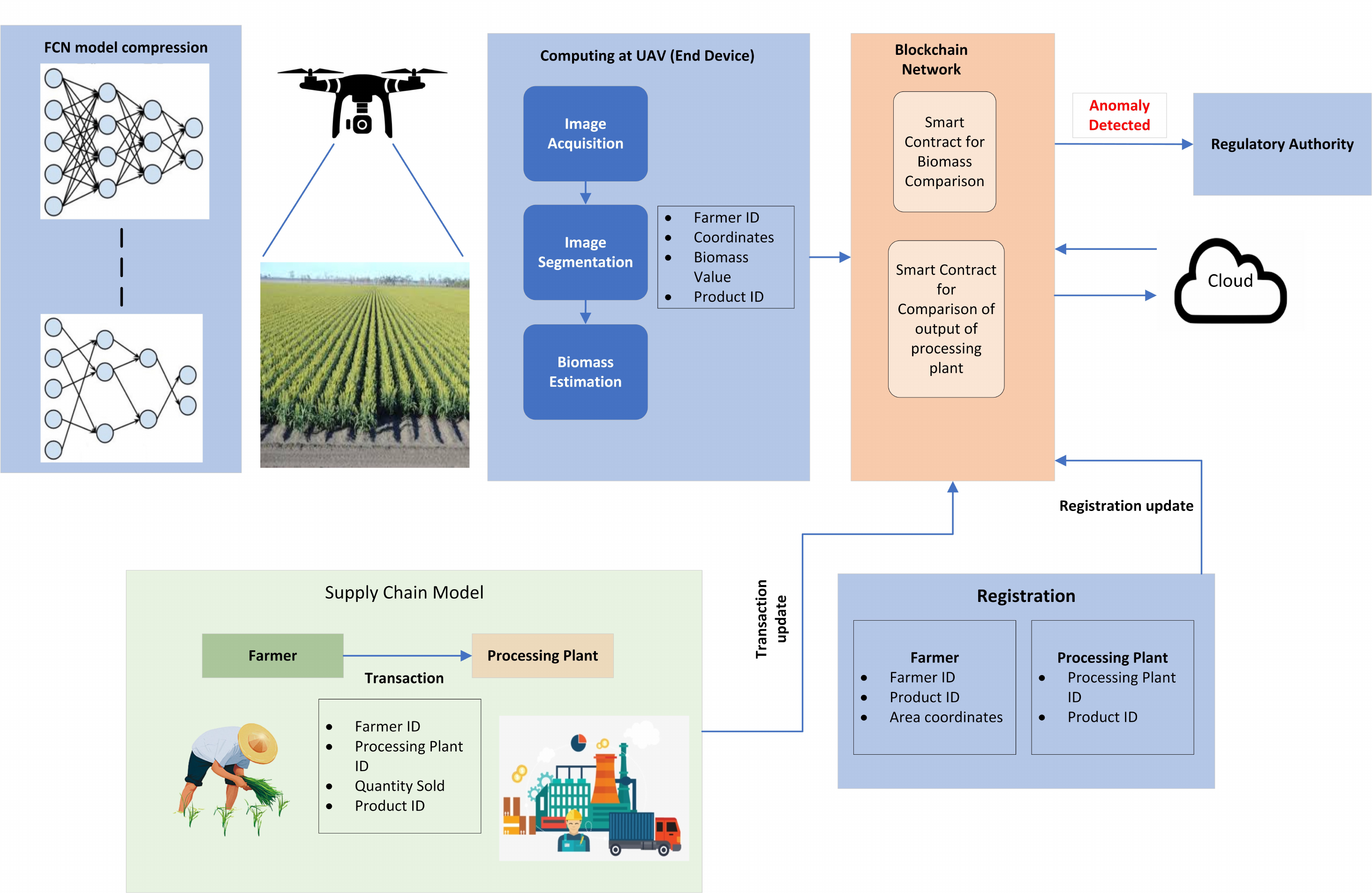}
\caption{Proposed AI and blockchain based agricultural supply-chain management network}
\centering\label{flow}
\end{figure*}

The motivation behind utilizing blockchain for supply-chain is to ensure transparency in information management. The system model is summarised in the flowchart shown in Fig. \ref{flow}. A detailed discussion of the flow chart is provided in Section \ref{systemmodel}. The rest of the paper is organized as follows. In section II, we present the motivation behind different aspects of our model. In section III, we present our system model, followed by proposed approach in Section IV, and section V presents the experimental setup and results, followed by conclusions and future directions in Section VI.

\section{ Motivation and Contribution}
In this section, we discuss the motivation behind different aspects of the proposed model. Specifically, in \ref{uavmot}, we present a brief overview of the role of UAVs for 6G networks and motivate their use in agri-food supply-chain management. In subsection \ref{edgemot}, we describe the significance of on-device approach over cloud-centric approach in the proposed use-case. \ref{blockchain} discusses the significance of blockchain to ensure traceability in food supply-chain. Lastly, section \ref{contributions} highlights the contributions of this work.

\subsection{UAVs and 6G networks} \label{uavmot}

UAVs have found a wide use in several applications ranging from security to agriculture, especially in the context of providing ubiquitous connectivity for 6G networks. Global mobile traffic has seen a mushroom growth in recent years, which builds the case for exploring new technologies such as UAVs to assist the traditional networks. UAVs are particularly helpful in realizing the concept of Heterogeneous networks (HetNets), where several small cells are deployed to enhance the network capacity. The main feature of UAVs that makes them an ideal solution for adhoc networks is their agility, allowing a quick and flexible deployment \cite{li}. Hence, they can be particularly useful for emergency networks. It is envisioned that UAVs will become a part of the wider terrestrial-air integrated network. However, this integration will also lead to challenges such as allocation of resources, reliability, security and path planning \cite{cao}.

Our aim is to exploit the UAVs for agri-food supply-chain management by capturing and processing the images of the agri fields. The main motivations behind employing a UAV for agri-food supply-chain management comprises of the following focal points: i) the ease of movement of UAVs makes them ideal for on-demand deployment to capture images of a field; ii), the UAVs possess the capability to process and transmit the information to the cloud; iii), UAVs can help in covering large area and can help in gathering images from different angles; iv), UAVs can be operated from a central location to introduce transparency for agri-food supply-chain management; v), UAVs can dynamically select a light weight AI-based solution that can reduce the bandwidth requirement and conserve resources; vi), UAVs can provide timely updates about the production to the supply-chain, thereby reducing the processing costs for large volumes of centralized data.

\subsection{On-Device vs Cloud-centric AI} \label{edgemot}
Deep Neural Networks (DNNs) as a subclass of AI algorithms have evolved over the past several years due to availability and accessibility of i) Big Data, ii) hardware accelerators, and iii) open-source software platforms to train and optimize them \cite{zhang01}. The superior performance of DNNs depends on their ability to extract high-level representations from raw data generated by Internet of Things (IoTs) and several other devices. Since DNNs are able to process a large amount of data, they require huge computational power to execute the floating point operations per second (FLOPS) inside their layers. A typical DNN is made up of multiple convolutional and fully connected layers stacked on top of each other, with thousands of nodes/filters interconnected in each layer that account for the millions of parameters and gigaFLOPS (GFLOPS) \cite{canziani}. Therefore, for training or inference of these DNNs, traditional IoT applications rely on the powerful capabilities of cloud data centres and high performance computers (HPCs). Generally, the raw data generated by devices is moved to the remote cloud data centre for processing and the decision is taken on the cloud once inferences are made. Although this status-quo approach alleviates the end devices’ load, the energy consumption, end-to-end latency and privacy issues occurring from wireless transmission are inevitable. On the other hand, pushing computations from cloud to end device is an emerging solution to the latency, energy, and privacy bottlenecks \cite{wang1}. In this approach, the DNN model is served on the end device that means data is being generated and processed at the same physical location. For instance, as shown in Fig. \ref{flow}, our approach motivates the on-device computation in smart agricultural application by computing the biomass from images on the UAV, instead of sending them to the cloud. Our approach ensures the privacy and low latency by restricting the raw data and computations within the boundaries of the farm. However, the low power end devices can not afford the computational complexity of DNNs. This pushed researchers to compress the DNN models so that they can be embedded into low power end devices \cite{han01}. 

There are various methods to make DNNs light-weight such as filter pruning, weights quantization, and knowledge distillation. Among them, filter pruning is simple, faster and efficient as it reduces redundant filters from the convolutional layers that accounts for most of the FLOPS \cite{xie01}. In this approach, the filters of each convolutional layer are ranked based on a certain scoring function such as average percentage of zeros, or  \textit{l}\textsubscript{1}-norm. These scoring functions are used to estimate the importance of filters towards the accuracy of the task, and then only the top-m ranked filters are kept intact for fine-tuning \cite{han01}. This single stage pruning approach results in the pruned and fine-tuned model with lower computational complexity for deployment on end devices. However, in the proposed scenario, we can not rely on a model with fixed complexity due to dynamic resource demands of the UAV. Thus, we motivate the scheme of providing a family of models with various resource-accuracy characteristics, so that the most efficient model can be selected based on the requirements of a certain IoT application.

\subsection{Blockchain} \label{blockchain}
The global agri-food supply-chain is a naturally dynamic structure that has evolved since the time of hunter-gatherers through subsistence agriculture. It is now a globalized environment with many moving pieces that make it much more complex. One of the biggest challenge includes a lack of cooperation among players due to individualistic mindsets and skewed views, as well as a lack of accountability, which may contribute to food security issues \cite{premanandh2013horse}. Such problems can be tackled by introducing blockchain technology in agriculture. From customer desire for transparency and more knowledge about the food they purchase to record-keeping and food integrity concerns, this exciting emerging technology can hold the key.

\subsection{Contributions}\label{contributions}
The contributions of this work are summarized as follows:
\begin{itemize}
    \item We propose a novel architecture for agricultural supply-chain management based on UAVs empowered by AI for on-device biomass estimation of the crops.
    \item In contrast to traditional biomass extraction techniques, we leverage the powerful capabilities of FCN model to provide pixel-wise label maps for biomass estimation. 
    \item We motivate the utilization of blockchain to keep track of crop's provenance and transactions (between farmers and processing plants) for ensuring traceability and transparency in the supply-chain network.
    \item Lastly, to operate an AI algorithm reliably over a 6G enabled dynamic UAV network, we provide a model selection approach using iterative model compression which generates multiple task-specific AI models with various complexities and accuracy trade-offs. As opposed to single light-weight model approaches, our results show trade-off among a variety of models to enable flexible AI on UAVs by selecting the desired model.
\end{itemize}
It is to be noted that our focus in this paper is not to propose a combination of AI and blockchain, but to propose an architecture based on on-device AI and blockchain for the novel application under consideration to address a serious issue in agriculture scenario. 
\section {System Model} \label{systemmodel}
\begin{figure*}[h]
\includegraphics[width=1\textwidth]{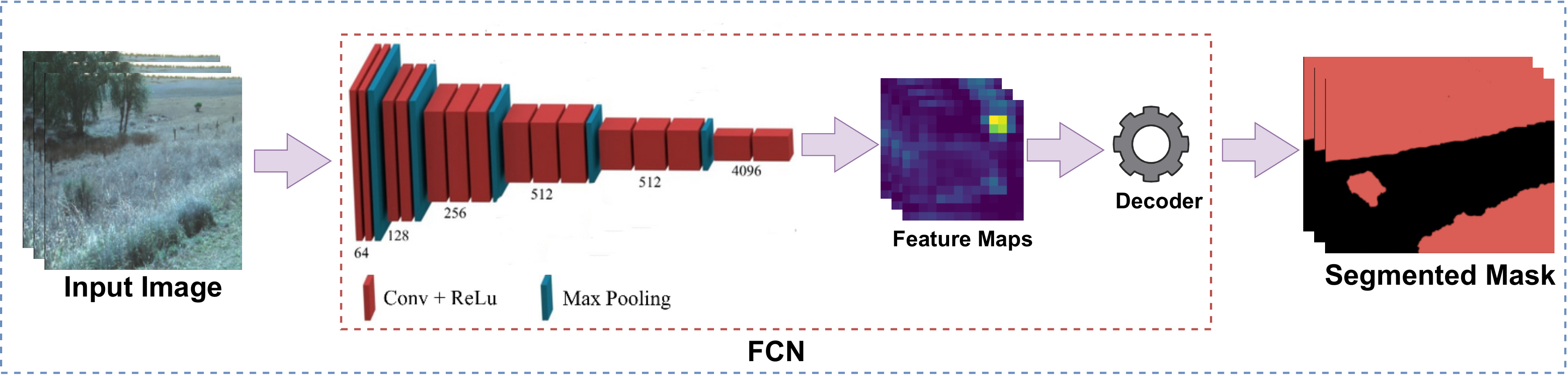}
\caption{FCN Architecture for semantic segmentation of Biomass Pixels}
\label{fcn}
\end{figure*}

The system model flow diagram is shown in Fig. \ref{flow}. In the proposed architecture, we assume that the UAV acts as an end device and collects images of the fields. The purpose of the proposed architecture is to collect the image data using UAVs, process it and transmit it to the cloud. The cloud collects the information provided by the UAVs and any transactions between a farmer and a processing plant are also uploaded to the cloud. The diagram also shows two smart contracts; First, for the biomass comparison by tracking the provenance of crops and second to compare the amount of raw crops which has been supplied to the processing plant and the prepared product provided by the processing plant to the market. This traceability helps in mitigating the risks of black marketing of the raw material as well as the final product at both farmers and industrial levels. In case of any transaction between the farmer and the processing plant, for example, if a farmer sells sugarcane to the processing plant, then the information will be updated through a blockchain network at the cloud. Each transaction generates information such as the farmer (seller) ID, the processing plant (buyer) ID, the quantity sold and the product ID. The processing at the UAV also involves the biomass comparison with updates provided by the transactions. In case of any anomaly between the biomass estimated value and the quantity updated by the farmer-processing plant transaction, a flag is generated and sent to the regulatory authority (e.g. government agency). Moreover, the outputs of the processing plant are also updated over a blockchain network. The aim is to check if the output of the processing plant corresponds to the input. Similar to the previous case, a flag is generated in case of any anomalies between the processing plant output and the projected output. The purpose of the proposed architecture is to introduce transparency in the supply-chain management.

\section{Proposed Approach based on AI and Blockchain}
Based on the above system model, this section demonstrate the AI based on-device biomass estimation, and blockchain based smart contract. 

\subsection{On-Device Biomass Estimation}
In real practice, the most common approach to calculate biomass is to conduct field surveys and visually observe the height of grasses using pre-defined criteria \cite{zhang02}. Although this process provides accurate results but it is labour intensive and time consuming as it involves human efforts. On the other hand, traditional image processing approaches rely highly on domain expertise and manual feature engineering on images to estimate biomass. However, deep learning techniques learn useful features automatically from the images with the help of convolutional filters \cite{long}. During training of CNNs, convolutional filters play a vital role in replacing the manual feature extractors. For this reason, they require minimal human intervention and provide state-of-the-art results. We approach the task of biomass estimation using semantic segmentation of biomass pixels in the images using FCN. Fig. \ref{fcn} shows the proposed FCN which takes an input image of arbitrary size and performs downsampling and upsampling using convolutional and transposed convolutional layers respectively to make pixel-wise predictions. The encoder part (downsampling layers) produce the class activation maps (CAMs) to increase the field of view (FOV) for the decoder part over input. Unlike traditional approaches, where images captured by end devices are forwarded to the cloud for biomass estimation, we propose on-UAV computation where UAV is responsible for both image acquisition and biomass estimation tasks. This approach not only reduces the latency and communication cost, but also improves privacy as data does not go out of the farm. Once an image is captured by UAV, it will be given to FCN which will perform semantic segmentation. Given the pixel-wise class labels of the image, the biomass will be calculated as the percentage of pixels identified as biomass. Lastly, instead of sending the images to the cloud, UAV will only update the estimated biomass to the blockchain along with the latitude and longitude of the respective farm land. 

\subsubsection{Model Compression} The complexity of CNNs is usually measured by the number of FLOPs and amount of memory they require for execution.  Thereby, the base FCN model which is used for on-device biomass estimation has approximately 125 GFLOPS, and consumes 513 megabytes (MBs) of memory when deployed \cite{canziani}.  This thus generates the corresponding need for memory requirements. Although modern UAVs are capable of handling significant amounts of computations and storage, it is not feasible to execute the raw model given the energy efficiency requirements. We employ filter pruning as a technique to compress the base model by removing the redundant filters from convolutional layers at a loss of small accuracy. For each filter \textit{F}, we compute its \textit{l}\textsubscript{1}-norm  (\norm{F}\textsubscript{1}) to rank its importance in a layer. The argument is that the filters with lower \textit{l}\textsubscript{1}-norm essentially generate very small activations, thus the corresponding filters can be pruned out. The pruning ratio determines the number of filters to be removed from each layer, and the model is fine-tuned on the rest of filters. The purpose of choosing \textit{l}\textsubscript{1}-norm over other scoring criteria is to reduce the extra computations required during the pruning process \cite{zhang01}. 

\subsubsection{Dynamic Model Selection} When performing on-UAV inference, the key challenges posed by UAV includes i) processing the incoming data with minimal delay to make critical decisions, and ii) consuming low energy in order to navigate smoothly and maximise flight time \cite{wang1}. Thus, to overcome these challenges, we propose a dynamic model selection strategy using iterative pruning. It is obvious from the discussion of model compression that accuracy can be traded off with lower complexity of compressed models. Therefore, our approach provides a roadmap of models with different accuracy, complexity, and energy trade offs. The required model can be fetched on-the-fly given the dynamic resource and accuracy requirements. For instance, if the processing capacity of UAV depletes during the flight, then the existing model can be replaced with a model having lower computational cost to avoid any failure.

\subsection{Smart Contract - Blockchain}

The main aim of employing blockchain in the proposed framework is to ensure the transparency in the supply-chain network comprised of farmers and processing plants. Moreover, regulatory authorities perform audits more often and trusting third party auditors is not a viable solution. Therefore, we ensure verifiable auditability with the help of blockchain networks. Firstly, farmers and processing plants are registered on the blockchain network and are assigned a unique blockchain address. This blockchain address can be used to see the historical transactions made on the blockchain network. Farmers are registered along with their unique ID, the crop type and latitude and longitude of their crop field. Similarly, processing plants are registered along with the unique IDs and product type. Once they are registered, they are assigned a unique blockchain address and they record all their trades on the blockchain network. Recording all the data will be advantageous for farmers as they will get good return for their crops. Moreover, the availability of different processing plants will provide farmers with the option to select the one with better rates. Similarly, processing plants can easily locate farmers with different quantities and types of crops. 

Once the entities are registered, they will start making transactions on the blockchain network regarding their trades. In our proposed blockchain based architecture, we are using two smart contracts, i.e., one to ensure the traceability and auditability of farmers and other for processing plant. UAVs over the crop fields are going to assist the smart contract for biomass comparison. UAVs will capture and analyse pictures and store this information on the blockchain network. This information will help smart contract to make a comparison if the farmers harvested crops. Moreover, if they are harvested, did farmers record the trade on the blockchain network? UAV assisted images will ensure transparency for the trades. The smart contract for biomass comparison will fetch the previously stored information on the blockchain network and compare it with the recent captured image by UAV. If the biomass data is different, smart contracts will emit an anomaly notification to the regulatory authority. Thus, the smart contract functionality will ensure transparency as well as auditability of the farmers. 

Similarly, we propose another smart contract in order to ensure auditability of processing plants, i.e., if they are utilizing all the crops to make products. Processing plants will report their products along with their quantity on the blockchain network. This stored information will be compared with the input crops quantity recorded by the farmers. If the information shows any discrepancy, a notification alert will be emitted to the regulatory authority. Therefore, the blockchain network will ensure the auditability of the processing plants and enforce them to do their duty honestly. 

The updating mechanism inside smart-contract simply involves updation of the estimated biomass value.  Smart-contract involves limited number of mathematical computations and the implementation complexity is thus reasonable.  However, it would  induce  an  additional  delay,  which  we  intend  to  investigate  in  our  future work.   Moreover,  it  must  be  noted  that  the  complexity  is  also  affected  by  the sampling size of the field area.  The smaller the sampling size the higher is the complexity as the computations have to be repeated more frequently and need to be updated via blockchain.
\begin{figure*}[h]
\includegraphics[width=1\textwidth]{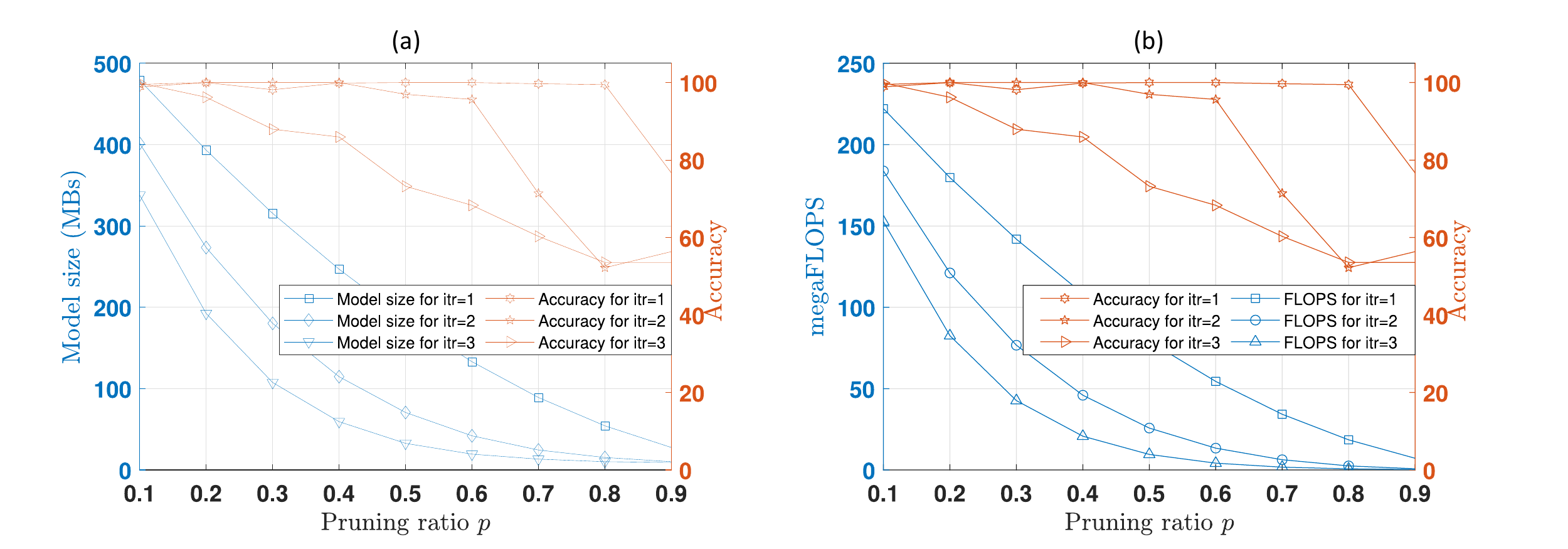}
\caption{(a) shows trade-off between accuracy and model size over different pruning ratios \textit{p} and iterations (itr), and (b) shows trade-off between accuracy and FLOPs over different pruning ratios \textit{p} and iterations (itr)}
\label{fig3}
\end{figure*}

\section{Experiments and results}
In this section, we provide an overview of our experiments and analyse the results achieved from iterative pruning technique. We show how iterative pruning can assist UAV to adaptively select a DNN model for on device semantic segmentation. The aim of our experiments is to explore multiple variants of a single application-specific DNN model with different computational complexities so that a suitable model can be picked up. 

\subsection{Training Baseline}

In order to run through the pruning experiments, we initially train a baseline DNN semantic segmentation model which is FCN. For training FCN, we use pre-trained weights of VGG-16 model on ImageNet dataset to initialize the parameters of the encoder part of FCN, while the parameters of decoder part are initialised randomly. The network is trained using Keras deep learning API with Tensorflow as backend on Nvidia Tesla K20 GPU. We fine tune the network on a dataset obtained from \cite{zhang02} after performing data augmentation techniques such as scaling, contrast transformation, mirroring, and rotation. On setting a learning rate of 10\textsuperscript{-2} with SGD optimiser, the model minimised the pixel-wise loss to a sufficient level on 128 epochs.

\subsection{Iterative pruning} 
The base model is not suitable for deployment directly on the UAV due to the resource and computational constraints. On the other hand, deploying a predetermined compressed model on the device does not satisfy the dynamic requirements of UAV. Therefore, our work addresses this challenge by using iterative pruning on pre-trained base model to create a roadmap of descendant models with various complexity-accuracy trade-offs as illustrated in Fig. \ref{fig3} (a) and (b). Instead of one stage - single pruning technique where the model is pruned for once with a very high pruning rate and then retrained, we use an iterative approach which is more effective and fast. In other words, if 30\% reduction in computational complexity is required, it is better to prune the model upto some iterations without retraining instead of pruning 30\% at once and again retraining the model. For this reason, we have explored the model with various iterations and pruning ratios so that the probability of choosing the right model increases. We perform experiments by pruning the model upto iteration (itr)=3, and for each iteration we vary the pruning ratio \textit{p} from 0.1 to 0.9 which gives us in total 27 variants of the base model. It is interesting to note that we achieved 98\% reduction in model size (Fig. \ref{fig3} (a)) and 99\% reduction in FLOPs (Fig. \ref{fig3} (b)) on itr=3 and p=0.9 but at the cost of losing 46\% accuracy. This scale of reduction is momentous when accuracy is not the concern for the application but low latency and minimal energy consumption is required. However, in our proposed framework, we not only have to take care of resource consumption but also accuracy, so given the threshold of accuracy and computational complexity, UAV can fetch the appropriate model from the local edge server taking full advantage of 6G. Therefore, referring to Fig. \ref{fig3} (a) and (b), as can be seen that with a small pruning ratio \textit{p}=0.2, the accuracy loss is very minimal upto 3\% at itr=3 as compared to single iteration itr=1. On the other hand, we achieve 54\% reduction in FLOPS, and 51\% reduction in model size at itr=3 and \textit{p}=0.2 as compared to itr=1 and \textit{p}=0.2. Thus, iterative pruning on the same ratio \textit{p} produces a much smaller model with minimal accuracy loss as compared to pruning only once. Moreover, there is no need to retrain the model because UAV needs to replace the model as quickly as possible in order to minimise the effect on flight time and battery. 
\begin{figure*}[h]
\includegraphics[width=1\textwidth]{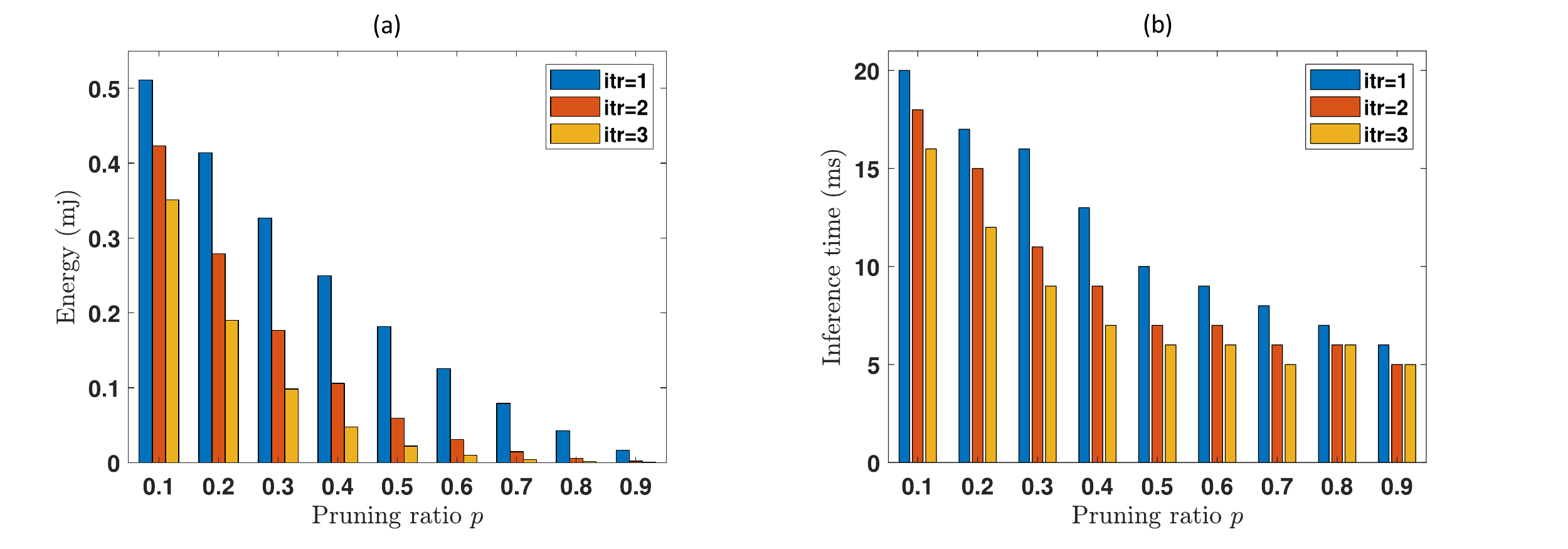}
\caption{(a) shows depletion of energy consumption over different pruning ratios \textit{p} and iterations (itr), and (b) shows depletion of inference time over different pruning ratios \textit{p} and iterations (itr)}
\label{fig4}
\end{figure*}
We also analyse the feasibility of the model in terms of the energy consumption and inference time as shown in Fig. \ref{fig4} (a) and (b) respectively. Inference time is measured as a delay in output required by the model to produce segmentation output. While the energy consumption of the model is based on the arithmetic operations and data access on the device. We assume that each 32-bit FLOP consumes 2.3 pJ, so energy required by arithmetic operations can be calculated by product of FLOP count and energy consumption of a single FLOP. Regarding the data access, we assume that retrieving 1MB of data from DRAM consumes 640 pJ, so for each model, it can be calculated as a product of model size and energy required to access each MB i.e. 640 pJ. The results shown in Fig. \ref{fig4} (a) and (b) are consistent with Fig. \ref{fig3} (a) and (b), because energy efficiency and inference time are essentially dependent on model size and FLOP counts \cite{xie01}. Thus, to efficiently manage the battery power and flight time of UAV, these two parameters can also be used as a selection criteria. Lastly, generating models with flexible resource-accuracy trade-offs at the runtime assist mobile vision systems like UAV to tackle dynamic resource needs by fetching the desired model. 

\begin{table}[ht]
\centering
	\caption{Comparison with state-of-the-art approaches on FCN model. $\downarrow$ denotes the reduction in percentage with respect to the original model. }
\begin{tabular}{|c|c|c|c|c|c|c|}
\hline
\textbf{Approach}    & \textbf{Accuracy} $\downarrow$  & \textbf{FLOPs} $\downarrow$   & \textbf{Memory}  $\downarrow$   \\ \hline
Mixed-Pruning \cite{mixed} &     5.09 \%  & 31.21 \% & 28.89 \%    \\ \hline
Network Slimming \cite{liu2017learning}  & 8.81 \% & 31.47 \% & 29.53 \% \\ \hline 
Proposed Approach   & 6.18 \% & 31.28 \% & 29.48 \% \\ \hline 
\end{tabular}
\label{table3}
\end{table}

\subsubsection{Comparative analysis}
For the sake of comparison, we have performed comparative analysis by reproducing a few state-of-the-art approaches on model pruning. We used same simulation parameters as mentioned in the original papers. In the Table \ref{table3}, we show the drop in accuracy, FLOPs and memory after pruning FCN model using the Mixed-pruning \cite{mixed}, network slimming \cite{liu2017learning}, and the proposed approach. Our approach losses 6.18\% of accuracy which is lesser than the network slimming which dropped 8.81\%. In contrast, mixed-pruning losses relatively lower accuracy as compared to ours, however, this approach is based on multi-stage process which introduces additional complexity in the compression scheme. Since our approach simply evaluates the \textit{l}\textsubscript{1}-norm of the filters which does not put additional burden, thus it outweighs the drop in accuracy as compared to mixed-pruning.

\begin{figure}[!h]
\centering
\includegraphics[width=0.5\textwidth]{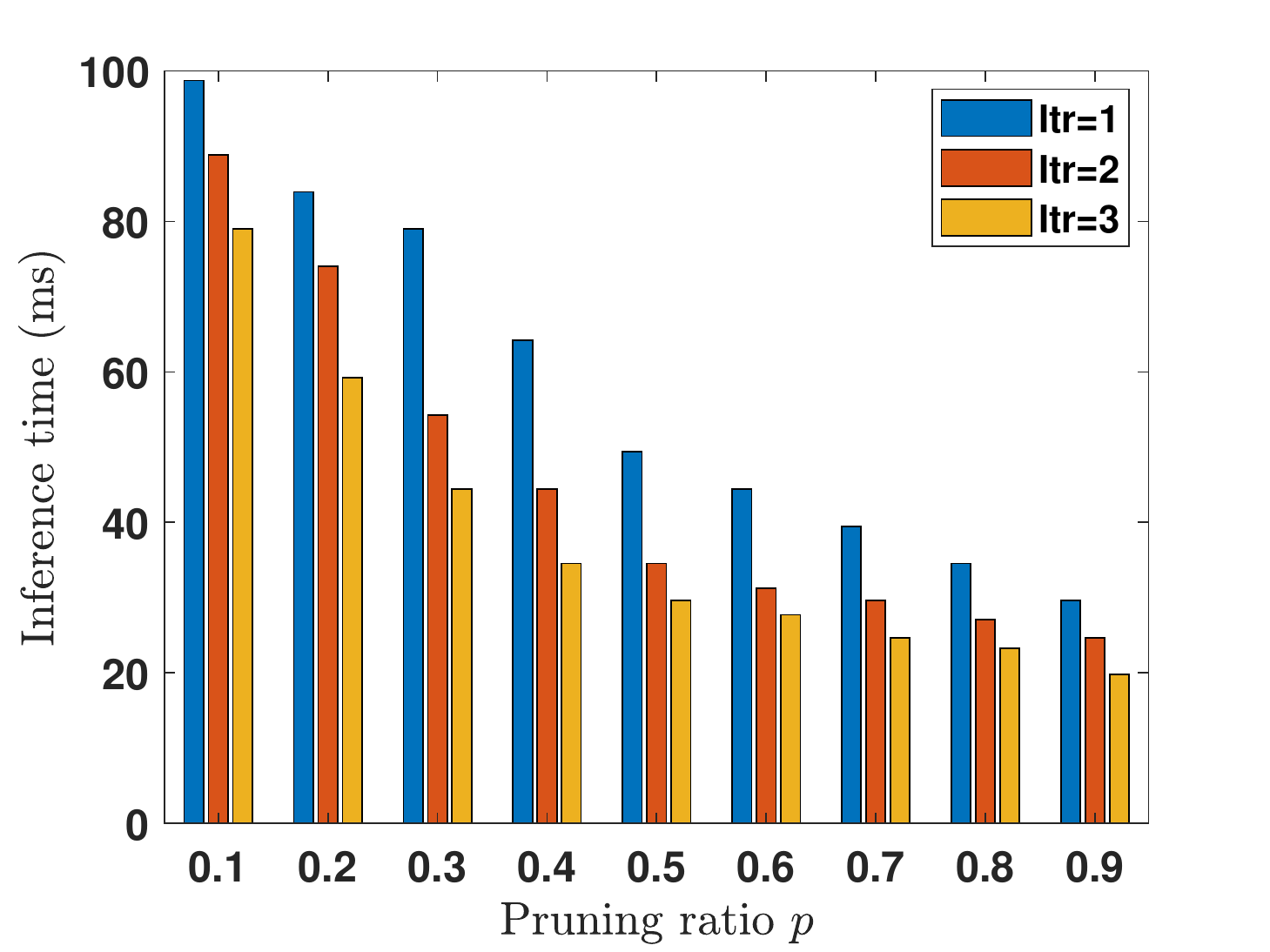}
\caption{Depletion of inference time over different pruning ratios \textit{p} and iterations (itr)}
\label{time2}
\end{figure}
\subsubsection{Usability in UAV scenario} 
We also evaluate the performance of iterative pruning on FCN model for the usability in UAV scenario. Since UAVs are usually resource-constrained, thus we evaluate our proposed approach on a resource-constrained device for the sake of simulations. We measure the latency of the FCN model on an OpenStack virtual machine (VM) with 2 CPUs, 4GB of RAM, and 10GB of the hard disk. We execute the FCN model on 100 random images, and show the average latency in Fig. \ref{time2}.  This latency indicates the delay incurred by FCN model in making an inference on an image while execution on a resource-constrained device. The unpruned model had approximately 107 ms of latency, while using the proposed pruning approach, the delay can be minimised with different pruning ratio and iterations as shown in Fig. \ref{time2}. This is significant for delay sensitive applications, for instance, using pruning ratio of 0.5 over a few iterations, we can achieve approximately 50\% reduction in latency. \\

\textit{Remark}: It is to be noted that for the sake of this paper, we have not considered the effects of the mobility, which in reality will affect the battery power and communication link of the UAV with the cloud. We aim to consider these effects in our future work. 

\section{conclusion and future works}
In this work, we motivate the utilization of UAVs for agriculture supply-chain management  by capturing images of the fields and estimating biomass through UAVs. Our proposed system model is based on the combination of blockchain and on-device AI. The aim of the proposed model is to ensure transparency and traceability. We proposed on-device processing of FCN for semantic segmentation of biomass pixels from the images captured by the UAV. Moreover, to deal with the dynamic resource management on UAV, we provide an iterative compression strategy which generates a road-map of models with resource-accuracy trade-offs. Instead of single compressed model which could cause failure during a UAV flight, we show in our results that with multiple iterations over different pruning ratios we can get multiple models for a specific task. Thereby, UAV can update the models with respect to changing resource requirements during the flight in order to be fault tolerant. 

In this paragraph, we highlight the limitations of our proposed architecture and provide a few future research directions to improve the proposed architecture. Firstly, the utilisation of intrusion prevention system with the proposed architecture is one of the future directions of this work. The intrusion prevention can help to secure the network against any unwarranted nodes (such as UAVs) and lead to an enhanced security. Secondly, privacy preservation mechanism can be utilized with the proposed architecture, which can add anonymity and persuade those users that are concerned about privacy to participate in the network. Moreover, battery power is one of the major constraints for sustaining the UAV based network. Therefore, the utilisation of a hybrid energy model that utilizes alternate energy sources such as solar can help in enhancing the network sustainability. The selection of the AI-based model can be done dynamically based on the available energy resources.

\section*{Acknowledgement}
This  research  was  supported  by  Science  Foundation  Ireland  and  the  Department  of  Agriculture,  Food  and Marine on behalf of the Government of Ireland VistaMilk research centre under the grant 16/RC/3835.
\ifCLASSOPTIONcaptionsoff
  \newpage
\fi
\bibliographystyle{IEEEtran}
\bibliography{mybibfile}

\end{document}